\documentclass{article}

\usepackage{amsmath, amssymb, amsthm, mathtools}
\usepackage{algorithm}
\usepackage{algpseudocode}
\usepackage{enumitem}
\usepackage{booktabs}
\usepackage[table]{xcolor}
\usepackage{pifont}
\usepackage{bm}
\usepackage{makecell}

\usepackage{wrapfig}
\usepackage{comment}
\usepackage{url}

\theoremstyle{plain}
\usepackage{hyperref}
\usepackage{cleveref}
\usepackage[utf8]{inputenc} 
\usepackage[T1]{fontenc}    
\usepackage{hyperref}       
\usepackage{url}            
\usepackage{booktabs}       
\usepackage{amsfonts}       
\usepackage{nicefrac}       
\usepackage{microtype}      
\usepackage{xcolor}         
\usepackage{multirow}
 \usepackage[preprint]{neurips_2026}

\newcommand{\method}{attention-state memory}
\newcommand{\Method}{Attention-state memory}
\newcommand{\cache}{memory}
\newcommand{\Cache}{Memory}
\newcommand{\Sufficiency}{Sufficiency}
\newcommand{\Composability}{Composability}

\newcommand{\attnop}{\operatorname{attn}}

\newcommand{\mergeop}{\operatorname{merge}}
\newcommand{\blendalpha}{\alpha}              

\theoremstyle{plain}

\theoremstyle{definition}

\theoremstyle{remark}

\newcommand{\R}{\mathbb{R}}

\title{Context Memorization for \\Efficient Long Context Generation}

\author{%
  Yasuyuki Okoshi\textsuperscript{1,2} \quad
  Hao (Mark) Chen\textsuperscript{2} \quad
  Guanxi Lu\textsuperscript{2} \quad \\
  {\bfseries Hongxiang Fan\textsuperscript{2} \quad
  Masato Motomura\textsuperscript{1} \quad
  Daichi Fujiki \textsuperscript{1}} \\
  \textsuperscript{1}Institute of Science Tokyo, Japan \quad
  \textsuperscript{2}Imperial College London, UK
}

\begin{document}

\maketitle

\begin{abstract}
Modern large language model (LLM) applications increasingly rely on long conditioning prefixes to control model behavior at inference time.
While prefix-augmented inference is effective, it incurs two structural limitations: \textit{i)} the prefix's influence fades as generation proceeds, and \textit{ii)} attention computation over the prefix scales linearly with its length.
Existing approaches either keep the prefix in attention while compressing it, or internalize it into model parameters through gradient-based training. The former still attends to the prefix at inference, while the latter is training-intensive and ill-suited to prefix updates.
To address these issues, we propose \method{}, a training-free approach that externalizes the prefix into a lightweight, lookup-based memory of precomputed attention states between prefix and query tokens. 
On ManyICLBench with LLaMA-3.1-8B, our method improves accuracy over in-context learning at 1K–8K memory budgets while reducing attention latency by 1.36× at 8K,  and surpasses full-attention RAG performance on NBA benchmark using only 20\% of its memory footprint.
Our code is available at \url{https://github.com/yasu0001/AttentionMemory}.
\end{abstract}

\vspace{-2ex}
\section{Introduction}
\label{sec:intro}
\vspace{-1ex}

From in-context learning~\citep{brown2020language, agarwal2024many} to external knowledge sources~\citep{lewis2020retrieval,chan2025don}, and agentic instructions~\citep{schick2023toolformer, yao2022react}, modern large language model (LLM) applications increasingly rely on long conditioning contexts (i.e., prefixes) to guide the behavior of LLMs during inference time.
While these prefix-augmented approaches improve model performance, they introduce two structural costs.
\textbf{The first is prefix decay:} as generation proceeds, the model's attention is distributed across tokens, decaying the influence of the prefix on model behavior~\cite{li2024measuringdrift,zhang2026tsubasa}, especially in long-context scenarios.
\textbf{The second is inference inefficiency:} as the prefix length increases, attention over the prefix imposes latency and memory overhead that scales linearly with its length on both prefill and every decode step~\cite{yang2025ape}, and prefix caching~\cite{kwon2023efficient,zheng2024sglang,jin2025ragcache}, though it amortizes prefill, still incurs substantial memory consumption.
This bottleneck is also prominent in deployed agentic systems: Anthropic reports that Claude Code is built around prompt caching (a form of prefix caching) to reduce latency and cost~\citep{anthropic2026promptcaching}, underscoring the need for methods that go beyond amortizing prefill and reduce the memory cost of prefix reuse.

Another line of research avoids re-attending to the prefix at inference time by internalizing prefix-conditioned behavior into model or adapter parameters, either through per-prefix fine-tuning (i.e., context distillation~\cite{snell2022learning, kujanpaa2024efficient, upadhayaya2024efficient, shin2025generative, asawa2026sieve}) or through a hypernetwork that maps prefixes to parameters in a single forward pass~\cite{charakorn2025text,charakorn2026doc}.
While eliminating attention on the prefix at inference time, these approaches inherit the cost of gradient-based training, making them slow, memory-intensive, and ill-suited to prefix updates.
On the other hand, hypernetwork based approaches~\cite{charakorn2025text,charakorn2026doc} only partially address this issue, as the hypernetwork itself requires training on billions of tokens.

To address these limitations, we propose a novel approach to eliminate inference-time attention over the prefix by retrieving precomputed attention states.
Rather than \textbf{internalizing} the prefix into model parameters through gradient-based training, we \textbf{externalize} it through forward-only computation, producing a lightweight, lookup-based memory.
Our approach offers three key advantages. 
First, it avoids the expense of gradient-based training, since the memory is built through forward-only computation.
Second, it removes the cost of attending to the prefix: lookup cost scales logarithmically with memory size, which is a hyperparameter independent of prefix length.
Third, the memory is decoupled from self-attention by retrieval, so its influence is less likely to decay as attention is drawn to generated tokens.

Concretely, \method{} constructs a memory of attention outputs between prefix and query tokens, then retrieves them at inference time.
The construction proceeds by running forward passes over a set of representative queries, collecting their attention outputs over the prefix, and clustering them into centroids.
At inference time, an incoming query retrieves the closest centroid and merges it with its self-attention.
By the online-softmax identity~\cite{rabe2021self,dao2022flashattention}, this merge process itself is lossless, recovering the attention output without attending to the prefix.

We evaluate on ManyICLBench~\cite{zou2025many} and RuleArena~\cite{zhou2025rulearena} to validate our \cache{} in both in-context learning (ICL) and retrieval-augmented generations (RAG) using LLaMA 3.1-8B~\cite{grattafiori2024llama}.
Our approach achieves downstream performance comparable to full attention scenarios while removing the prefix attention.
Specifically, on ManyICLBench, \method{} improves accuracy over in-context learning at 1K--8K memory budgets while reducing attention latency by $1.36\times$ at 8K.
For RAG, our method surpasses full-attention RAG performance on the NBA benchmark using only $20\%$ of its memory footprint. Therefore, our contributions are:
\begin{itemize}
    \vspace{-2ex}
    \item We propose \method{}, a training-free, lookup-based attention-state dictionary that externalizes long prefixes into a compact memory.
    \item We extend the online-softmax identity from efficient attention computation to cross-query prefix reuse.
    \item Experiments on ICL and RAG benchmarks demonstrate that \method{} matches or exceeds full-attention performance while reducing prefix attention cost.
    \vspace{-2ex}
\end{itemize}

\begin{figure}
    \centering
    \includegraphics[width=0.9\linewidth]{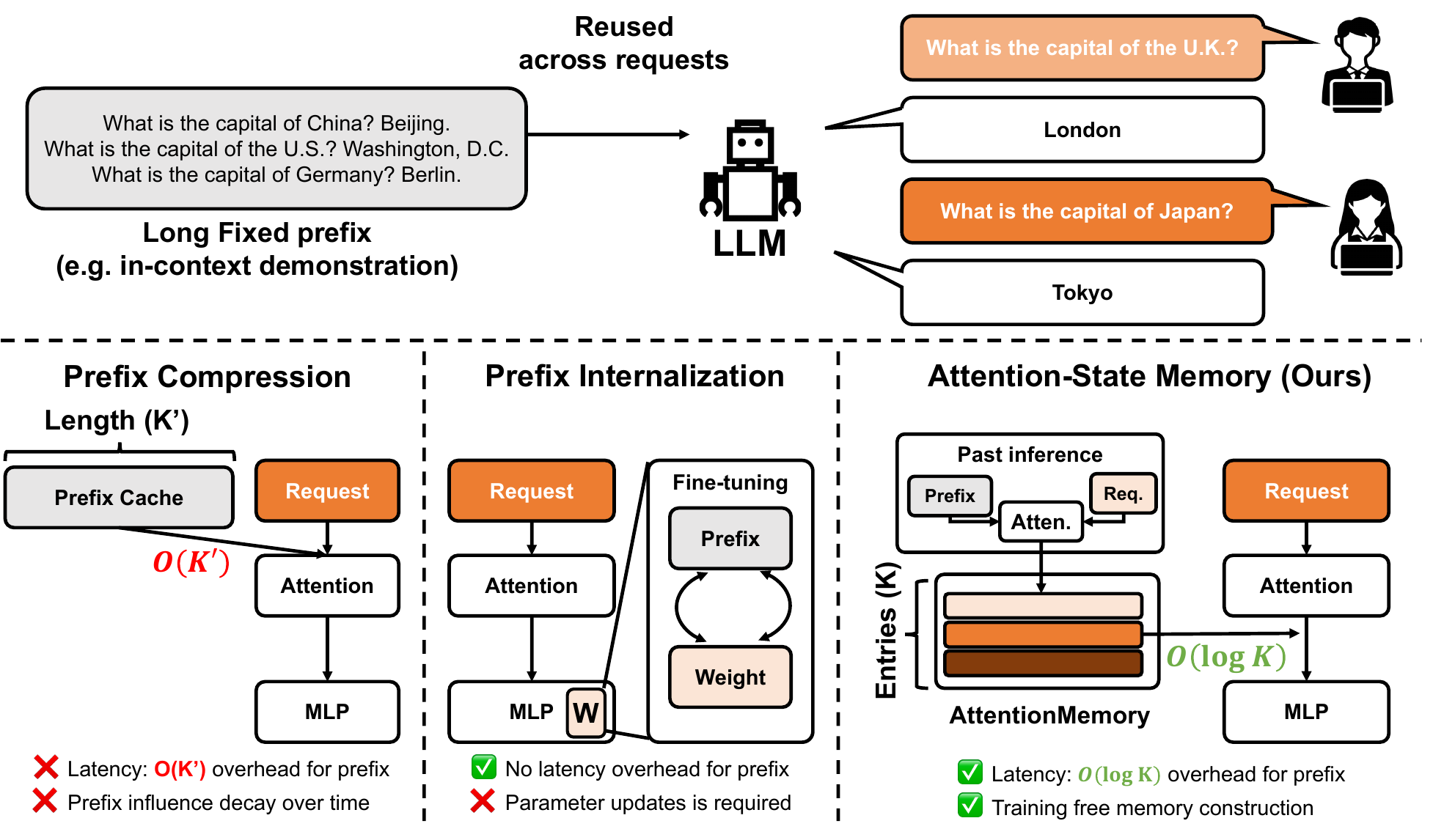}
    \caption{Comparison of three approaches for handling a long fixed prefix that is reused across many user requests (top).
    In-context learning attends to the prefix per step ($O(L)$); 
    fine-tuning absorbs it into parameters via gradient training; 
    Attention-State Memory (ours) externalizes it into a $K$-entry lookup ($O(\log K)$) built by forward-only inference.}
    \label{fig:introduction}
\end{figure}

\vspace{-1ex}
\section{Background}
\vspace{-1ex}
\label{sec:background}

\vspace{-1ex}
\subsection{Related Work}\label{sec:bg:related_works}
\vspace{-1ex}
Existing work that reduces the prefix cost can be broadly categorized into two families based on whether the prefix is removed from inference-time attention: (i) prefix internalization and (ii) prefix compression.

\textbf{Prefix internalization.} \quad
A line of work removes the prefix from attention at inference time by encoding it into model parameters.
The research directions can be categorized into two approaches based on whether these parameters are produced through gradient descent or meta-network.
Context distillation~\citep{snell2022learning, kujanpaa2024efficient, upadhayaya2024efficient, shin2025generative, asawa2026sieve, zhang2026tsubasa} fine-tunes the model on each prefix so that its outputs without the prefix match those obtained with it, while hypernetwork-based approaches~\citep{charakorn2025text, charakorn2026doc} amortize this per-prefix cost by mapping prefixes to low-rank parameters in a single forward pass.
Both avoid prefix decay and eliminate prefix overhead at inference, but require gradient-based training, which is resource-intensive, sensitive to hyperparameters.

\textbf{Prefix compression.} \quad
A separate line of work keeps the prefix inside attention while reducing its size.
Prompt compression shortens the prefix at the token level: hard methods~\cite{jiang2023llmlingua, li2023selectivecompressing,pan2024llmlingua} prune low-information tokens to produce a shorter natural-language prefix, while soft methods~\cite{mu2023learninggist,chevalier2023autocompressor,ge2023contextice} encode the prefix into a small number of continuous tokens through a trained encoder-decoder pipeline.
Query-agnostic KV cache compression~\cite{kim2025kvzip,song2026csattention} operates at a lower level by evicting or selecting entries inside the KV cache, and reuses it across queries.
Both avoid the per-prefix training cost of internalization, as the compressed prefix is constructed without gradient backpropagation.
However, both leave attention over the compressed prefix at inference time, so the cost of attending to the prefix is reduced rather than removed, and the influence of prefix is remains subject to decay as attention is drawn to generated tokens~\cite{li2024measuringdrift,zhang2026tsubasa}.

Overall, to our knowledge, no prior method simultaneously provides prefix-length-independent decoding latency, training-free prefix construction, and no auxiliary models. 
Methods that keep the prefix inside attention preserve flexibility but pay per-query attention, while methods that move it into parameters eliminate that attention but require gradient updates to incorporate or refresh a prefix.

\vspace{-1ex}
\subsection{Online-Softmax Identity}
\label{sec:bg:online-softmax}
\vspace{-1ex}
Our approach builds on the online-softmax identity~\cite{rabe2021self}, which has also been applied in efficient attention implementations such as FlashAttention~\cite{dao2022flashattention, dao2023flashattention, shah2024flashattention} and MAC-Attention~\cite{yao2026mac}.
Attention over a concatenated key block can be losslessly decomposed into attention over its sub-blocks, where the combination weights are determined by the dot-product of the query–key scores within each sub-block.
Let $q \in \R^{d_h}$ be a query vector with per-head dimension $d_h$, and let $K = [K_1,\dots,K_B] \in \mathbb{R}^{L\times d_h}$, $V = [V_1,\dots,V_B] \in \mathbb{R}^{L\times d_h}$ be the keys and values over a prefix of length $L$, partitioned into $B$ disjoint blocks.
Attention over $K, V$ can be decomposed into:
\begin{equation}
  \attnop(q, K, V) \;=\; \sum_{b=1}^{B} \blendalpha_b \, \attnop(q, K_b, V_b),
  \qquad
  \blendalpha_b \;=\; \frac{\sum_{k\in K_b}\exp(qk/\sqrt{d_h})}{\sum_{b'}\sum_{k\in K_b'}\exp(qk/\sqrt{d_h})}.
  \label{eq:lse-blend}
\end{equation}
For simplicity, we denote $a_b(q) = \attnop(q, K_b, V_b)$ and $Z_b(q)= \sum_{k\in K_b}\exp(qk/\sqrt{d_h})$ in the remaining paper.

\paragraph{Implications.}
The attention decomposition implies two opportunities for the proposal.
First, \textbf{\Sufficiency}: for a given query, storing $(a_b(q), Z_b(q))$ is sufficient to reconstruct the block's contribution to attention without loss. 
In this case, the original keys and values are no longer needed.
In this paper, we refer to $(a_b(q), Z_b(q))$ as \textbf{attention state}.

Second, \textbf{\Composability}: two attention states for the same query over disjoint key--value blocks $(K_1, V_1)$ and $(K_2, V_2)$ can be merged into a single attention state via the online-softmax update.
We define the merge operator $\mergeop(\cdot, \cdot)$ by
\begin{align}
    \mergeop((a_A(q), Z_A(q)),(a_B(q), Z_B(q))) \;&\triangleq\; \Bigl(\frac{Z_A(q)\, a_A(q) + Z_B(q)\, a_B(q)}{Z_A(q) + Z_B(q)},\; Z_A(q) + Z_B(q)\Bigr), \label{eq:combine}
\end{align}
which recovers exactly the attention state over the concatenated block:
\begin{equation}
    \mergeop((a_A(q), Z_A(q)),(a_B(q), Z_B(q))) = (a_{[A, B]}(q), Z_{[A, B]}(q)).
    \label{eq:combine-equiv}
\end{equation}
Here, we represent $[A, B]$ as the concatenation of two blocks.
Applying this rule repeatedly, we can compute attention states independently for each block and merge them at inference time, recovering the attention over the concatenated blocks—equivalent to parallel encoding~\cite{ratner2023parallel, yang2025ape}.

Together, \Sufficiency{} and \Composability{} suggest a new way to handle long fixed prefixes: rather than attending to the prefix at inference or internalizing it into model parameters, we can externalize it into a precomputed dictionary of attention states.
We realize this idea in \Cref{sec:method}.
\vspace{-1ex}
\section{Attention-State Memory}
\vspace{-1ex}
\label{sec:method}

\begin{figure}
    \centering
    \includegraphics[width=1.0\linewidth]{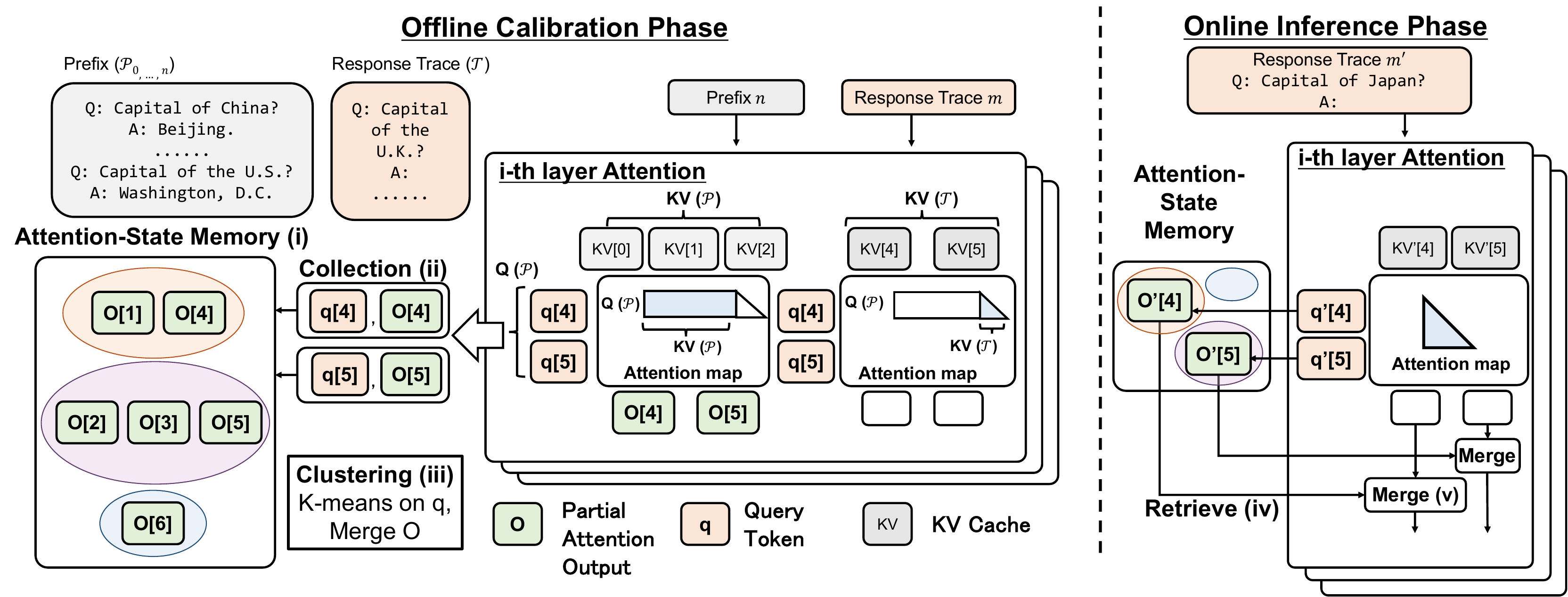}
    \caption{Overview of \method{}. }
    \label{fig:method_overview}
\end{figure}

\subsection{Key Insight from Implications}
\vspace{-1ex}

The two properties of the attention decomposition~(\Cref{sec:bg:online-softmax}) imply that prefix attention can be externalized into a query-based dictionary, which can be constructed and updated entirely through forward passes.
\textbf{\Sufficiency{}} enables lossless recovery via lookup: since attention states fully determine prefix attention, precomputing them for a fixed query set allows prefix attention to be recovered through a dictionary lookup at inference time.
\textbf{\Composability{}} enables forward-only construction and update: the memory can be assembled from independently encoded prefix chunks and extended with new prefixes through a single forward pass.

Together, these properties define an idealized memory bank of a query-indexed dictionary of $(a_p(q), Z_p(q))$.
In practice, storing one entry per possible query is infeasible, so we approximate the idealized dictionary by representative entries $K$ obtained by clustering on a calibration set.

\vspace{-1ex}
\subsection{Overview of Attention-State Memory}
\vspace{-1ex}
\Method{} (ASM) is a per-layer dictionary of attention states $(a, Z)$, indexed by representative query vectors and shared across queries through clustering.
\Cref{fig:method_overview} provides an overview.
At construction~(\Cref{fig:method_overview}, left), we run a forward pass over a concatenated set of prefix and response traces.
For every tokens in response traces, we collect its query vector together with its attention state $(a, Z)$ over the prefix, then apply clustering to the query vectors to compress these triples into fixed entries per layer.
During inference~(\Cref{fig:method_overview}, right), a query searches for the memory entry with the highest similarity, and retrieves a pre-computed attention state $(a, Z)$.
Then, these values are merged into the query's self-attention, without the need to compute attention to the prefix.

\vspace{-1ex}
\subsection{Memory Bank}
\vspace{-1ex}
\label{sec:method:object}

A key feature of ASM is a per-layer dictionary of pre-computed attention state (\Cref{fig:method_overview} (i)).
For each layer $i \in \{0, ..., N_{\text{layer}}{-}1\}$ the \cache{} $C^{(i)}$ consists of $K$ entries:
\begin{equation}
  \mathcal{C}^{(i)} \;=\; \Big\{\big(\bar{q}_k^{(i)},\, \bar{a}_k^{(i)},\, \bar{Z}_k^{(i)}\big)\Big\}_{k=1}^{K}
  \quad\text{with}\quad
  \bar{q}_k^{(i)} \in \mathbb{R}^{d}, \;
  \bar{a}_k^{(i)} \in \mathbb{R}^{H \times d_h}, \;
  \bar{Z}_k^{(i)} \in \mathbb{R}^{H},
  \label{eq:cache-object}
\end{equation}
where $H$ is the number of attention head,
$\bar{q}_k^{(i)}$ is the lookup key with dimension $d{=}Hd_{h}$, and $(\bar{a}_k^{(i)}, \bar{Z}_k^{(i)})$ is the compressed attention state.
For simplicity, this formulation assumes the standard multi-head attention, where each head maintains its own KV cache.
We explain the extension to grouped-query attention (GQA)~\cite{ainslie2023gqa} in \Cref{sec:method:gqa}.

We use the query as the lookup key following the standard of KV cache compression approaches~\cite{zhang2023h2o, li2024snapkv, hooper2025squeezed}.
The key assumption behind our method is that the attention output from a similar set of tokens would produce a close representation~\cite{yao2026mac}.

In the following sections, we explain how to construct \cache{} and how to retrieve it.

\subsection{Offline Calibration Phase}
\vspace{-1ex}
\label{sec:method:centroid}

We construct the ASM from a prefix set $\mathcal{P}{=}\{\mathbf{p}\}_{j=1}^{N}$ and a response trace set $\mathcal{T} {=}\{\mathbf{t}_j\}_{j=1}^{M}$ in a offline-manner.
The prefix set contains contextual information to the model, such as in-context examples, task instructions, or retrieved documents.
Each response trace contains a user prompt and a response.
\Cache{} construction proceeds in two phases: collection and clustering.
\vspace{-1ex}
\paragraph{Collection phase (\Cref{fig:method_overview} (ii)).}
For each prefix-trace pair $(\mathbf{p}, \mathbf{t})\in \mathcal{P}\times \mathcal{T}$, we obtain the KV cache at each layer by running a forward pass over the concatenated sequence $[\mathbf{p}, \mathbf{t}]$.
For each query in a response trace $q\in Q^{(i)}(\mathbf{t})$, we record the prefix attention state over the prefix $(a^{(i)}_{\mathbf{p}}(q), Z^{(i)}_{\mathbf{p}}(q))$.
Aggregating across all traces produces a set $\mathcal{P}^{(i)}= \{(q, a^{(i)}_{\mathbf{p}}(q), Z^{(i)}_{\mathbf{p}}(q)): q\in Q^{(i)}(t), \mathbf{t}\in\mathcal{T} \}$ at each layer, of size $|\mathcal{P}^{(i)}| = N\sum_{\mathbf{t}\in \mathcal{T}}|Q^{(i)}(\mathbf{t})|$ ($N$ times the total number of tokens across all traces).

While we use $N{=}1$ in most experiments, $N{>}1$ naturally arises when the prefix is chunked for efficient online calibration described below or when multiple documents are retrieved as in retrieval-augmented generation (RAG).

\vspace{-1ex}
\paragraph{Clustering phase (\Cref{fig:method_overview} (iii)).} We partition each $
\mathcal{P}^{(i)}$ into $K$ clusters via K-means on the query representations $\{q : (q, \cdot, \cdot) \in \mathcal{P}^{(i)}\}$, and aggregate each cluster into a single entry $(\bar{q}, \bar{a}^{(i)}, \bar{Z}^{(i)})$.
For the aggregation step, we propose attention-aware aggregation, which preserves the merge structure of attention in~\Cref{eq:combine-equiv}.
The centroid of each cluster $C_k$ is computed by merging its attention states using \Cref{eq:combine}:
\begin{equation}
    \bar{q}_k
      \;=\; \frac{1}{|\mathcal{C}_k|}
            \sum_{(q,\, \cdot,\, \cdot) \in \mathcal{C}_k} q,
    \qquad
    \bar{Z}_k
      \;=\; \frac{1}{|\mathcal{C}_k|}
            \sum_{(\cdot,\, Z,\, \cdot) \in \mathcal{C}_k} Z,
    \qquad
    \bar{a}_k
      \;=\; \frac{\displaystyle\sum_{(\cdot,\, Z,\, a) \in \mathcal{C}_k} Z \, a}
                  {\displaystyle\sum_{(\cdot,\, Z,\, \cdot) \in \mathcal{C}_k} Z}.
    \label{eq:centroid-finalize}
\end{equation}
We normalize $\bar{Z}_k$ by $|C_k|$ so that the centroid acts as an average rather than an unbounded merge, motivated by prior findings that combining many independently encoded contexts without normalization degrades performance due to attention scale mismatch~\cite{yang2025ape}.
\vspace{-1ex}
\paragraph{Efficient offline calibration.}
While constructing ASM requires only a forward pass, the peak GPU memory still scales linearly with prefix length.
This cost becomes prohibitive when the prefix spans tens of thousands of tokens, potentially limiting the practical applicability of ASM on memory-constrained devices.
To address this, we exploit the compositional structure of ASM: the $\mergeop$ operator in \Cref{eq:combine-equiv} exactly combines two $(a(q), Z(q))$ pairs from disjoint prefixes into a single pair that recovers the attention state of their concatenation, enabling parallel encoding~\cite{ratner2023parallel,yang2025ape} of long prefixes.
A long prefix can therefore be partitioned into chunks, encoded independently, and merged within the \cache{}.
For instance, a 16K-token prefix can be constructed from four independent 4K-token forward passes.

\vspace{-1ex}
\subsection{Online Inference Phase}\label{sec:method:inference}
\vspace{-1ex}
At inference time, the model takes only the user query as input and generates the response without attending to the prefix.
To incorporate the prefix representation, we retrieve the corresponding attention state from the memory and merge it into the attention between the user query.
This section explores each part in detail.
\vspace{-1ex}
\paragraph{Retrieval (\Cref{fig:method_overview} (iv)).}
The memory retrieval is performed independently for each layer and each user query token.
At layer $i$, the incoming query is used as the lookup key (the specific representation is discussed in below).
We find the nearest cluster centroid by cosine similarity following~\cite{matsushima_aqpim}:
\begin{equation}
  c^\star(q)
  \;=\;
  \arg\max_{k \in \{1,\dots,K\}} \;
  \cos\!\big(q,\, \bar{q}_k^{(i)}\big)
  \;=\;
  \arg\max_k \; \frac{\langle q,\, \bar{q}_k^{(i)} \rangle}
                    {\|q\|\,\|\bar{q}_k^{(i)}\|}.
  \label{eq:retrieve}
\end{equation}
Given $c^\star$, we retrieve the compressed attention state $(\bar{a}^{(i)}_{c^\star}, \bar{Z}^{(i)}_{c^\star})$ for use in the merge step.

\vspace{-1ex}
\paragraph{Merge (\Cref{fig:method_overview} (v)).}
For each query $q$ at layer $i$, we merge the retrieved attention state $(\bar{a}^{(i)}_{c^*}, \bar{Z}^{(i)}_{c^*})$ with the user query attention $(a(q), Z(q))$ computed from standard self-attention over the non-prefix tokens.
Following the merge structure in \Cref{eq:combine-equiv}, the merged attention output is:
\begin{equation}
 a_{\text{merge}}(q)
 \;=\;
 \frac{Z(q)}{Z(q) + \bar{Z}^{(i)}_{c^*}} \, a(q)
 \;+\;
 \frac{\bar{Z}^{(i)}_{c^*}}{Z(q) + \bar{Z}^{(i)}_{c^*}} \, \bar{a}^{(i)}_{c^*}.
 \label{eq:merge-blend}
\end{equation}
The merged output $a_{\text{merge}}(q)$ then proceeds through the rest of the attention block as usual.
\vspace{-1ex}
\paragraph{Memory lookup key.} \quad
The retrieval in \Cref{eq:retrieve} uses a query-side representation as the \cache{} lookup key.
The choice of representation determines which queries are grouped into the same cluster during construction, and which cluster is selected at inference time.
We consider two orthogonal design choices, the choice of RoPE handling and the choice of whitening, resulting in four configurations that we explore in the paper.

$\bullet$~\textbf{\textit{RoPE: pre-RoPE vs RoPE-unified}.}\quad
We consider two ways of constructing the query representation. 
The first uses the output of query projection before rotary embedding is applied. This representation does not depend on absolute position and captures purely semantic similarity. 
The second applies rotary embedding at a common virtual position across all queries. This captures both positional and semantic similarity.

$\bullet$~\textbf{\textit{Whitening}.} \quad
Independent of the RoPE choice, we optionally apply a whitening transform to the lookup key, following prior work that has shown whitening to improve cosine-similarity retrieval~\cite{su2021whitening,huang2021whiteningbert}.
In typical backbones, the variance of query projection outputs is uneven across dimensions, so cosine similarity becomes dominated by the high-variance dimensions rather than reflecting task-relevant signal.
We address this by applying
\begin{equation}
  q \;\gets\; \Sigma^{-1/2} q,
  \label{eq:whiten}
\end{equation}
where $\Sigma$ is the sample covariance of $Q$, computed on a random subsample of $\mathcal{T}$, independently for each layer and each attention head.

\textbf{Efficient online inference.} \quad
A linear lookup over all K entries takes $\mathcal{O}(K)$ time per query.
By indexing centroids hierarchically, this cost drops to $\mathcal{O}(\log K)$.
Hierarchical lookup~\cite{jegou2010product,johnson2019billion} decouples retrieval cost from the number of memory entries and allowing the memory to grow without proportionally increasing inference latency.
\vspace{-1ex}
\subsection{Attention-State Memory for GQA}
\vspace{-1ex}
\label{sec:method:gqa}
We now describe how ASM extends to grouped-query attention (GQA)~\citep{ainslie2023gqa}.
Under GQA, $G = H_q / H_{kv}$ query heads share a single KV head, where $H_q$ and $H_{kv}$ are the numbers of query and KV heads.
Since the $G$ query heads in a group attend to the same KV head, the resulting attention outputs depend on the same prefix keys and values. 
This redundancy means that we can store a single centroid per KV head, rather than per query head, without losing information.
Concretely, for each group in a layer, we form an aggregated query $q\in \mathbb{R}^{d'}$, which concatenates the per-head queries within the group. We then collect the aggregated queries and corresponding attention states across calibration data. Thus, the number of collected samples per group becomes $Gd_h / d'$ times larger than that of the standard multi-head attention.
We then cluster these collected samples to construct centroids, which can be done in two ways: clustering the aggregated queries independently per group, or jointly after concatenating them across groups. These two strategies trade per-group fidelity against centroid count and lookup cost.

\textbf {Memory footprint of attention-state memory.} \quad
During decode, standard GQA loads the entire KV cache, incurring prefix traffic of $2Hd_h + 2LGd_h$, where $2Hd_h$ covers the query load and output write, and $2LGd_h$ covers loading the keys and values over all $L$ prefix tokens.
ASM with $K$ entries retrieves $1$ entry per query, incurring traffic of $2Hd_h + KGd'$, where $2Hd_h$ covers the query load and the intermediate output write, and $KGd'$ covers loading the $K$ lookup keys (each of dimension $d'$) over the \cache{}.
When $K{=}L$, setting $d'{=}2d_h$ matches the prefix traffic of \method{} to that of standard GQA, and we adopt this as the default throughout our experiments.

\section{Experiments}\label{sec:experiment}
\vspace{-1ex}
\subsection{Experimental Settings}
\vspace{-1ex}
\paragraph{Benchmarks.}
We evaluate on two complementary scenarios that reflect the dominant uses of long prefixes: in-context learning (ICL) and retrieval-augmented generation (RAG).
For ICL, we use seven tasks from ManyICLBench~\cite{zou2025many}---five reported to show large many-shot gains in prior work~\cite{zou2025many} and two reasoning-oriented tasks in math and science.
For RAG, we use the NBA bench from RuleArena~\cite{zhou2025rulearena}, which provides $\sim$20K tokens of player-trade regulations.
We exclude other RuleArena tasks since baselines achieve near-zero accuracy even with full in-context rules.

\vspace{-1ex}
\paragraph{Attention-state memory construction.}

\Method{} (ASM) is constructed from each task's training split; for NBA bench, we use synthetic data following~\cite{asawa2026sieve}, filtering any sequences overlapping the test set.
The \cache{} is built from a 32K-token prefix for ICL and a 20K-token rulebook for NBA, with entry counts varied over $\{1, 2, 4, 8, 16\}$K.
Unless otherwise stated, we set $d'=2d_h$ so that per-entry memory footprint matches a single KV cache entry under standard GQA (\Cref{sec:method:gqa}).
The number of construction iterations is determined per task by training-data size (Appendix~\ref{app:iterations}), scaled by $1.25\times$ at 16K entries for clinc150 and banking77 where centroid convergence requires more iterations.
We sweep the four key modes and two centroid organizations and report the best validation configuration per task to understand design choices effects.

\vspace{-1ex}
\paragraph{Models and baselines.}
We evaluate our method on the instruction-tuned variant of LLaMA 3.1-8B~\cite{grattafiori2024llama}.
We compare against two baselines that represent the upper and lower bounds of long-prefix usage.
The first is full-context, in which the entire prefix is provided to the model at inference time.
The second is no-context, in which the model recieves only a brief task description in the prompt without any in-context examples or retrieved rules.
We additionally compare ASM against KVzip, a training-free KV cache compression method on the five ICL tasks.
We omit prompt compression from this comparison as it operates at the token level, a substantially coarser granularity than the KV-level compression of KVzip.
To ensure a fair comparison, we apply a uniform per-layer memory budget for both KVzip and our method.
KVzip is also compress 32K-token prefix of ICL.
We provide additional model evaluations in Appendix~\ref{sec:exp:model}.
\vspace{-1ex}
\subsection{Benchmarking on In-Context Learning Tasks}
\begin{figure}
    \centering
    \includegraphics[width=1.0\linewidth]{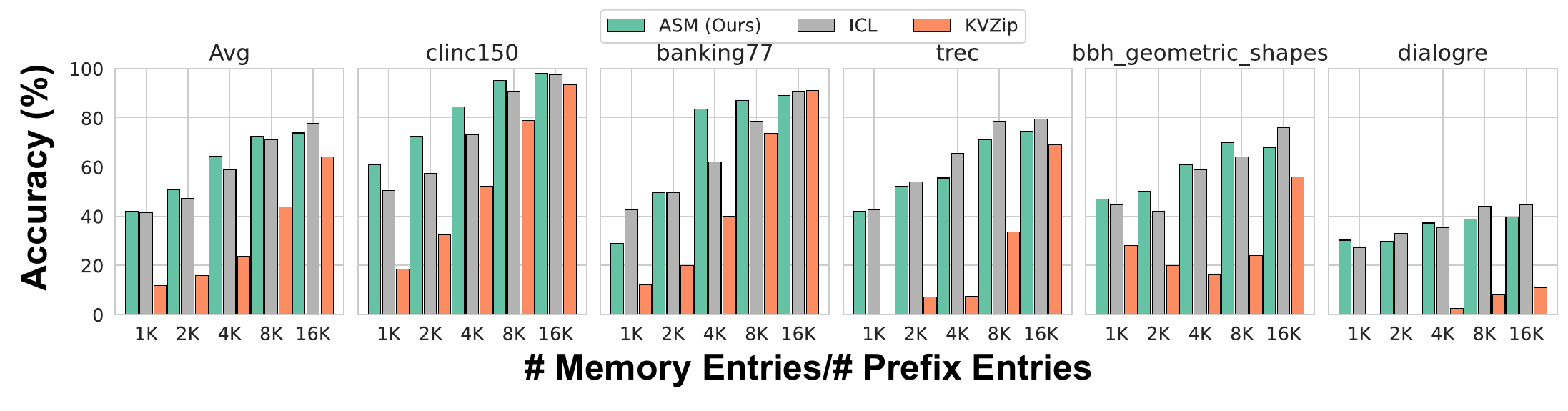}
    \caption{Accuracy of five benchmarks from ManyICL Bench. X-axis represents the number of memory entries (Ours) or the number of KV cache entries (ICL and KVzip). The same number of entries incurs the same memory footprint, as discussed in \Cref{sec:method:gqa}.}
    \label{fig:icl_results}
\end{figure}

\vspace{-1ex}
\Cref{fig:icl_results} compares ASM with ICL in varying memory entries on five subset tasks of ManyICL benchmark.
We also compare our method against KVZip, a KV cache compression method.
Across the five benchmarks, ASM outperforms or matches ICL across all memory budgets from 1K to 8K in average performance.
We also observe that ASM outperforms ICL up to 8K memory entries, and within 4K entries it also scales more efficiently.
This efficiency comes from the fact that each entry stores the attention output over the entire prefix.
As a result, \method{} can keep contextual information that ICL would lose when its context size is limited.
At the very small memory entries ($K{=}1$K), the gain becomes marginal as each entry has to summarize too many different training examples, and the average becomes too coarse to be useful.

The primary reason behind the performance crossover at 16K is insufficient training data for k-means centroid construction at larger bucket sizes.
Some benchmarks (e.g., bbh\_geometric\_shapes with only 150 training samples) do not provide enough data to populate a 16K codebook, preventing ASM from scaling effectively as the number of buckets grows.

KVZip underperforms both ASM and ICL across most settings, suggesting that KV cache compression collapses critical information such as label tokens in the ICL prompt.
In contrast, \method{} stores attention outputs that aggregate information over the entire prefix, preserving such critical signals regardless of the compression ratio.

\begin{table}[t]
\centering
\begin{minipage}[t]{0.49\linewidth}
    \centering
    \caption{Accuracy of ICL and ASM (ours) on reasoning benchmarks (math\_counting and gpqa\_cot) across different memory sizes. Bold marks the best accuracy across lengths for each method.}
    \begin{tabular}{l c c c c}
\toprule
\multirow{2}{*}{\# Entries} & \multicolumn{2}{c}{math\_counting} & \multicolumn{2}{c}{gpqa\_cot} \\
\cmidrule(lr){2-3} \cmidrule(lr){4-5}
 & ICL & \textbf{ASM} & ICL & \textbf{ASM} \\
\midrule
1K  & 23.9          & 24.0 & \textbf{28.3} & \textbf{28.3} \\
2K  & 22.0          & 23.0 & \textbf{28.3} & 24.2          \\
4K  & 25.0 & 22.5          & 27.3 & 27.8          \\
8K  & \textbf{26.0} & 24.5          & 24.2 & 23.7          \\
16K & 22.5          & \textbf{26.0} & 22.7          & 23.7 \\
\bottomrule
\end{tabular}
    \label{tab:reasoning-bucket-sweep}
\end{minipage}
\hfill
\begin{minipage}[t]{0.49\linewidth}
    \centering
    \caption{Accuracy on NBA benchmark in RuleArena. Bold marks the best accuracy across length for each method.}
    \begin{tabular}{l c c}
\toprule
Method & \# Entries & Accuracy \\
\midrule

Zero-shot  & 0K  & 21.2 \\
ICL  & 20K        & \textbf{24.1} \\
\midrule
\multirow{5}{*}{\textbf{ASM}}
  & 1K   & 19.4 \\
  & 2K   & 19.9 \\
  & 4K   & \textbf{25.5} \\
  & 8K   & 23.2 \\
  & 16K  & {19.4} \\
\bottomrule
\end{tabular}
    \label{tab:rag}
\end{minipage}
\end{table}

We further evaluate on two reasoning benchmarks, math and science, whose results are summarized in~\Cref{tab:reasoning-bucket-sweep}.
In these benchmarks, the ICL itself shows only marginal improvement on these tasks.
Since our \cache{} shares the same foundation as in-context learning, its accuracy tracks the baseline closely across bucket sizes from 1K to 16K.
However, ASM achieves similar performance to the ICL baseline on both tasks, indicating that \cache{} retains the prefix information effectively.

These results suggest that the benefit of ASM is closely tied to how ICL itself scales with the memory budget. On reasoning tasks, where ICL accuracy does not consistently improve with longer sequences, the gains of ASM are correspondingly modest.
\vspace{-1ex}
\subsection{Benchmarking on Retrieval-Augmented Generation}
\vspace{-1ex}
This section evaluates ASM on the NBA benchmark from RuleArena to analyze its effectiveness in RAG.
We compare ASM at varying entries against (i) a zero-shot baseline (no rulebook) and (ii) an ICL baseline (full 20K rulebook in context).
\Cref{tab:rag} reports exact-match accuracy between the reference source and the expected answer.
ASM surpasses the full-rulebook ICL baseline at its best setting ($K{=}4$K), using only 20\% of the memory footprint and without providing reference rules at inference time.
Notably, accuracy is non-monotonic in the number of entries and peaks at an intermediate codebook size, suggesting that the optimal number of entries is independent of the prefix length and should be tuned as a task-specific hyperparameter rather than simply set as large as possible.

\begin{figure}[t]
\centering
\begin{minipage}[t]{0.3\textwidth}
\centering
\includegraphics[width=1.0\textwidth]{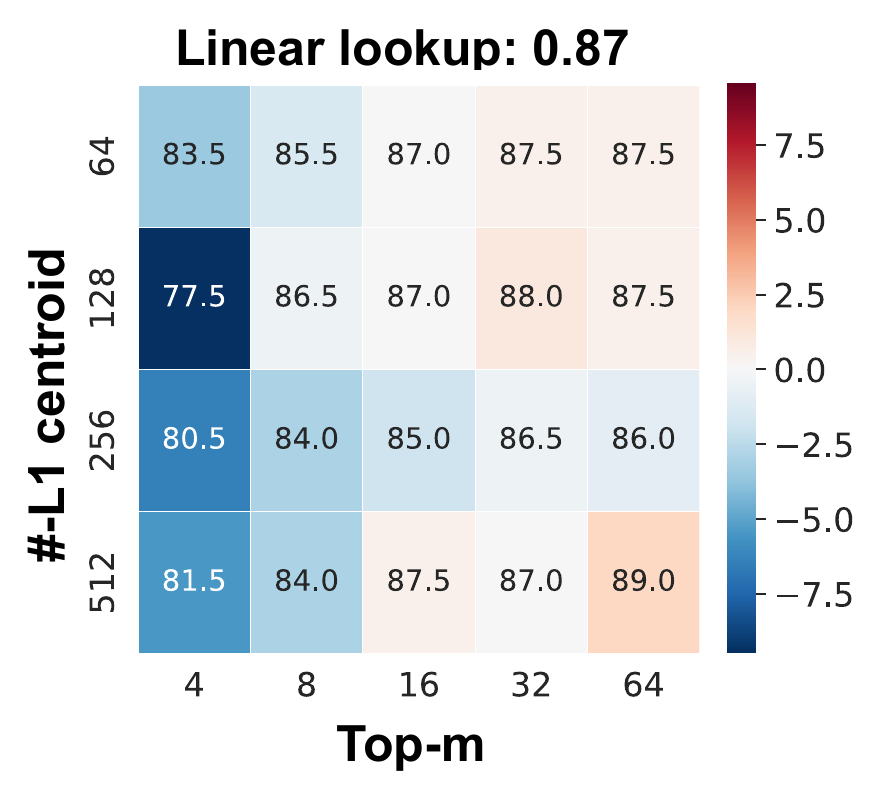}
\caption{Heatmap for hierarchical lookup. Blue indicates regions where hierarchical search underperforms linear lookup, while red indicates the opposite.}
\label{fig:hierarchical}
\end{minipage}
\hfill
\begin{minipage}[t]{0.68\textwidth}
\centering
\includegraphics[width=1.0\textwidth]{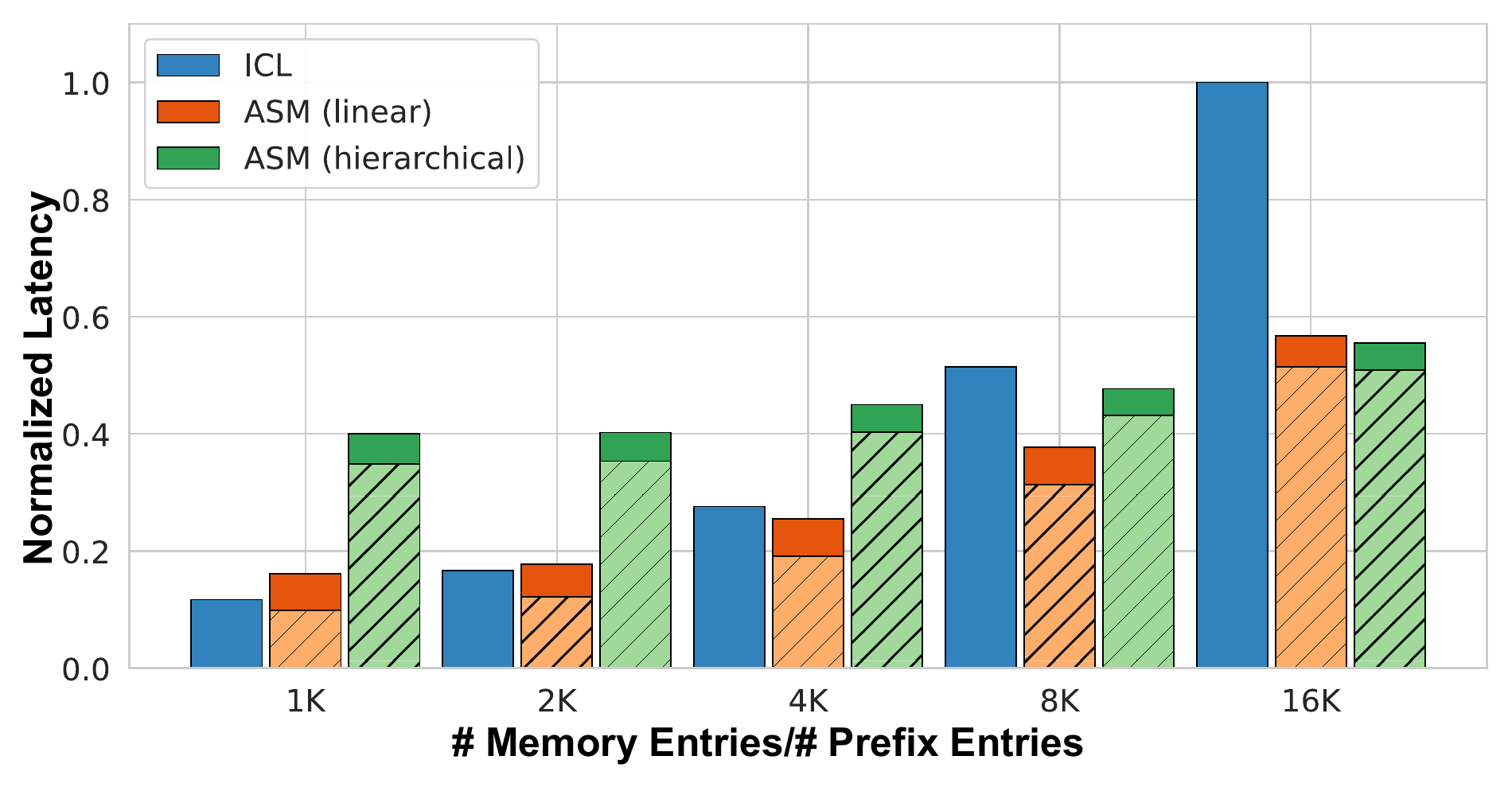}
\caption{Attention latency comparison between existing in-context learning and \method{} with linear lookup and hierarchical lookup. For hierarchical lookup, we first retrieve the top-16 first-level centroids, then perform a linear search over their associated second-level centroids to find the closest one.}
\label{fig:latency}
\end{minipage}%
\end{figure}

\begin{wraptable}{r}{0.45\textwidth}
\vspace{-3em}
\centering
\caption{Effect of chunked prefix construction on banking77 accuracy.}
\label{tab:bank77-curriculum}
\scalebox{0.85}{\begin{tabular}{c c c c}
\toprule
Chunk size &\# Chunks & Total iter   & Accuracy \\
\midrule
4K & 1           & $5{,}000$    & $61.5$  \\
16K & 1           & $1{,}250$    & $79.0$  \\
4K & 4            & $5{,}000$    & $78.5$  \\
\bottomrule
\end{tabular}
}
\end{wraptable}

\vspace{-1ex}
\subsection{Efficiency Analysis}
\vspace{-1ex}
\paragraph{Efficient offline calibration.}
We empirically validate the chunked construction enabled by $\mergeop$, where a long prefix is encoded as multiple independent forward passes over shorter chunks.
Table~\ref{tab:bank77-curriculum} reports accuracy on banking77 when constructing a \cache{} over a 16K-token prefix from four 4K-token chunks via $\mergeop$, and compares it against two reference points: a 16K baseline that processes the entire prefix in a single forward pass, and a 4K baseline that uses only a single 4K chunk.

The chunked construction closely matches the accuracy of the 16K baseline while never processing more than 4K tokens in a single forward pass.
Furthermore, given the same total number of tokens consumed across forward passes, the chunked construction achieves substantially higher accuracy than the 4K baseline, indicating that the additional chunks contribute complementary information rather than redundant computation.
Together, these results show that the compositional structure of \method{} enables long-prefix calibration with substantially reduced peak memory, while preserving the performance of full-prefix construction.
\vspace{-1em}
\paragraph{Efficient online inference.}
Our algorithm can reduce the retrieval cost from $\mathcal{O}(K)$ to $\mathcal{O}(\log K)$ by indexing centroids hierarchically.
\Cref{fig:hierarchical} validates this property by comparing linear lookup against hierarchical lookup at varying the number of first-level clusters expanded to the second (top-$m$) and the number of first-level centroids $n_\text{L1}$, for $K{=}8$K.
Hierarchical retrieval with top-$m \geq 16$ matches flat top-1 accuracy across most settings of $n_\text{L1}$.
We therefore adopt top-$m{=}16$ in the subsequent latency evaluation (\Cref{sec:exp:latency}), as it is the smallest value that achieves this and thus minimizes the retrieval cost.
\vspace{-2ex}
\subsection{Latency Analysis}\label{sec:exp:latency}
\vspace{-1ex}
We measure the average per-token latency of the attention block over the generation of 512 tokens, in order to analyze how attention cost scales with the memory budget.
To ensure a fair comparison, the x-axis represents a shared memory footprint as described in~\ref{sec:method:gqa}.
We compare three configurations: full attention, ASM with linear lookup, and ASM with hierarchical lookup.
Full experimental details are provided in Appendix~\ref{app:iterations}.

The ICL baseline shows approximately linear growth with prefix length, as each decoding step attends to the entire prefix.
In contrast, the retrieval overhead of hierarchical lookup does not scale linearly with the number of memory entries.
It incurs a relatively higher overhead at small memory budgets, but this overhead grows much more slowly than full attention as the budget increases.
By combining linear and hierarchical lookup into an optimized ASM configuration, our method becomes faster than full attention at around 4K memory entries and achieves a $1.8\times$ speedup at 16K entries.

These results demonstrate that, unlike standard self-attention where compute scales together with memory cost, \method{} suppresses the growth of inference time as the memory budget increases.

\vspace{-2ex}
\section{Conclusion}
\vspace{-2ex}
In this work, we address the problem of decoupling prefix knowledge from per-query attention computation, enabling prefixes to be reused across queries without repeated full attention over the entire context.
We propose \method{}, a training-free approach that externalizes the prefix into a lightweight, lookup-based memory of precomputed attention outputs between prefix and query tokens.
Across ICL and RAG benchmarks, \method{} achieves a favorable trade-off between accuracy and efficiency. On ICL, it surpasses full attention in the 1K--8K memory entries range on average while reducing attention latency by 1.36× at 8K. On RAG, it outperforms full attention with only 20\% of the memory footprint.
More broadly, we view this work as one step toward a broader research direction of externalizing LLM knowledge into compact, reusable representations beyond text and model parameters.

\newpage
\bibliographystyle{plain}
\bibliography{contents/99.reference}

\appendix
\section{Detailed Hyperparameters}\label{app:iterations}
\begin{table}[t]
\centering
\small
\caption{Lookup key configuration and centroid grouping strategy for LLaMA 3.1-8B. Lookup key configuration and grouping strategy is discussed in~\Cref{sec:method:inference} and \Cref{sec:method:gqa}, respectively.}
\label{tab:key-bucketing-per-task}
\setlength{\tabcolsep}{6pt}
\begin{tabular}{l c l l}
\toprule
Task & Iterations & Lookup key & Shared Centroid \\
\midrule
banking77                         & $5{,}000$ & pre-RoPE           & Individual \\
clinc150                          & $5{,}000$ & pre-PoPE           & Individual \\
trec                              & $5{,}000$ & pre-RoPE           & Shared \\
dialogre                          & $1{,}000$ & pre-RoPE           & Individual \\
bbh\_geometric\_shapes            & $150$ & post-RoPE       & Shared \\
math\_counting                    & $1{,}510$ & pre-RoPE           & Invididual \\
gpqa\_cot                         & $344$  & pre-RoPE + Whitening & Shared \\
nba                               & $6{,}000$ & pre-RoPE + Whitening & Shared \\
\bottomrule
\end{tabular}
\end{table}

\paragraph{Hyperparameters.}
\Cref{tab:key-bucketing-per-task} summarizes the per-task memory configuration used throughout the main experiments.
The number of construction iterations is determined by the size of the available training split for each task, which directly bounds how many distinct training examples can be observed during centroid updates.
For tasks with large training pools (banking77, clinc150, trec, nba), we use $5{,}000$--$6{,}000$ iterations, while tasks with limited training data (bbh\_geometric\_shapes, gpqa\_cot) are capped at the number of available examples.
The memory lookup key and sharing strategy are selected per task based on validation accuracy from the sweep, with pre-RoPE and invididual centroid emerging as the most common choice across both ICL and reasoning tasks.
Reasoning tasks which is sensitive tokenization (gpqa\_cot, nba) benefit from the whitening variant, which we attribute to the role of leading-space tokens in distinguishing answer choices.

\paragraph{Latency evaluation.}
We measure the latency of attention during decoding to directly compare the efficiency of our method against standard attention.
All measurements are conducted on a single NVIDIA RTX Ada 4500 GPU with batch size 1, a question length of 512 tokens, and a generation length of 100 tokens, while varying the prefix length or \cache{} entries from 1K to 16K.
Standard attention is computed with FlashAttention~\cite{dao2022flashattention}, and the key lookup in our method is implemented as a custom Triton~\cite{tillet2019triton} kernel.
We report latency normalized to the slowest attention baseline.

\paragraph{Compute resources.}
All experiments except for latency evaluations are conducted on a single NVIDIA H100 GPU.
Memory construction time depends on the number of iterations and the prefix length, with the longest configuration (NBA bench) taking approximately $1.5$ hours.
Evaluation time is dominated by token generation and similarly takes up to $1.5$ hours on NBA bench, the task with the longest generation length.
All other tasks complete substantially faster than these upper bounds.

\section{Generalizability of Attention-State Memory to Different Models}\label{sec:exp:model}
This section explores whether \cache{} transfers across architectures and scales.
To validate this, we evaluate \method{} on LLaMA-3.2-3B and Qwen3-8B, which respectively isolate the effect of model size and the effect of model family relative to LLaMA-3.1-8B used in our main experiments.
Evaluations are conducted on two benchmarks, banking77 and NBA benchmark, chosen as representatives of the ICL and RAG settings. The former is the ICL task with the largest gain from in-context examples in \Cref{fig:icl_results}, where the effect of caching is most clearly observable.
We exclude reasoning tasks, since \Cref{tab:reasoning-bucket-sweep} shows ICL provides no measurable benefit on them, making them uninformative for studying backbone dependence.

\Cref{tab:banking77-qwen3-vs-llama32} and \Cref{tab:nba-qwen3-vs-llama32} report the results on banking77 and NBA bench, respectively.
On banking77, ASM matches or exceeds the same-budget ICL baseline across memory sizes for both backbones, with the gap being most pronounced at smaller budgets and narrowing as the budget grows.
On NBA bench, ASM consistently outperforms both the full-rulebook ICL baseline and the zero-shot setting on both models, following the trend observed on Llama-3.1-8B in our main experiments.
We also note that the optimal memory size differs between the two backbones, which we attribute to differences in attention head geometry and grouping factor $c$ across architectures.
Overall, these results indicate that ASM generalizes across both model scale and model family without architecture-specific tuning.


\begin{table}[t]
\centering
\small
\caption{Banking77 accuracy on Qwen3-8B vs Llama-3.2-3B-Instruct. For each model we show the best cache configuration
(pre-RoPE, Individual) and its ICL baseline at the same prefix memory trafic. ASM row varies the number of memory entries; baseline row varies prefix tokens for ICL.}
\label{tab:banking77-qwen3-vs-llama32}
\setlength{\tabcolsep}{4pt}
\begin{tabular}{l l c c c c c c}
\toprule
Model & Setting & 1K & 2K & 4K & 8K & 16K \\
\midrule
\multirow{2}{*}{Qwen3-8B}
  & ICL
    & \textbf{50.0} & 50.0 & 66.0 & 82.0 & \textbf{89.5} \\
  & \textbf{ASM}
    & 32.0 & \textbf{57.0} & 71.5 & 81.0 & 83.5 \\
\midrule
\multirow{2}{*}{Llama-3.2-3B}
  & ICL
    & 39.5 & 44.0 & 60.5 & 76.0 & 85.5 \\
  & \textbf{ASM}
    & \textbf{48.0} & \textbf{67.5} & \textbf{77.5} & \textbf{84.0} & \textbf{85.5} \\
\bottomrule
\end{tabular}
\end{table}

\begin{table}[t]
\centering
\small
\caption{NBA benchmark accuracy on Qwen3-8B vs Llama-3.2-3B-Instruct.
Cache rows use Individual centroid with pre-RoPE key mode and
all-layer injection. \textbf{Bold} marks the per-row best.}
\label{tab:nba-qwen3-vs-llama32}
\setlength{\tabcolsep}{5pt}
\begin{tabular}{l l c c c c c c}
\toprule
Model & Setting & & 1K & 2K & 4K & 8K & 16K \\
\midrule
\multirow{3}{*}{Qwen3-8B}
  & Zero-shot (empty rulebook) & $30.6$ &\multicolumn{5}{l}{} \\
  & ICL, full rulebook         & $29.6$ & \multicolumn{5}{l}{} \\
  & Cache, group-concat        & & \textbf{31.5} & $27.8$ & $31.0$ & $26.9$ & $25.9$ \\
\midrule
\multirow{3}{*}{Llama-3.2-3B}
  & Zero-shot (empty rulebook) & $26.9$ &\multicolumn{5}{c}{} \\
  & ICL, full rulebook         & $26.9$ &\multicolumn{5}{c}{} \\
  & Cache, group-concat        & & $30.1$ & $31.5$ & $29.2$ & $31.5$ & $\mathbf{33.3}$ \\
\bottomrule
\end{tabular}
\end{table}

\section{Limitation}\label{sec:limitation}
Our method assumes that queries exhibit local structure, allowing a small set of cached centroids to faithfully approximate attention. This holds in the settings we target, such as ICL and RAG, where queries from a fixed task or persona tend to cluster.
However, the assumption can break down in scenarios with substantial prefix drift, such as long multi-turn conversations spanning casual, technical, and task-oriented exchanges, where query distributions become diffuse.
In such cases, query-dependent retrieval may need to be replaced with alternative lookup strategies tailored to non-stationary distributions.
Extending the externalization idea behind \method{} to such diverse workload is a promising direction for future work.


\end{document}